\documentclass[conference]{IEEEtran}
\IEEEoverridecommandlockouts
\usepackage{cite}
\usepackage{amsmath,amssymb,amsfonts}
\usepackage{algorithmic}
\usepackage{graphicx}
\usepackage{textcomp}
\usepackage{xcolor}
\usepackage{array}
\usepackage{booktabs}
\usepackage{multirow}
\usepackage{subcaption}
\usepackage{orcidlink}
\usepackage{authblk}
\usepackage{cite}

\hypersetup{
    colorlinks=true,
    urlcolor=magenta,      
}

\def\BibTeX{{\rm B\kern-.05em{\sc i\kern-.025em b}\kern-.08em
    T\kern-.1667em\lower.7ex\hbox{E}\kern-.125emX}}
\begin{document}

\title{GAIS: A Novel Approach to Instance Selection with Graph Attention Networks}

\author[1,2]{Zahiriddin Rustamov \orcidlink{0000-0003-4977-1781}}
\author[3]{Ayham Zaitouny}
\author[1]{Rafat Damseh}
\author[1,*]{Nazar Zaki}

\affil[1]{College of Information Technology, United Arab Emirates University, Al Ain, UAE}
\affil[2]{Department of Computer Science, KU Leuven}
\affil[3]{College of Science, United Arab Emirates University, Al Ain, UAE\authorcr{\tt nzaki@uaeu.ac.ae}}

\maketitle

\begin{abstract}
    Instance selection (IS) is a crucial technique in machine learning that aims to reduce dataset size while maintaining model performance. This paper introduces a novel method called Graph Attention-based Instance Selection (GAIS), which leverages Graph Attention Networks (GATs) to identify the most informative instances in a dataset. GAIS represents the data as a graph and uses GATs to learn node representations, enabling it to capture complex relationships between instances. The method processes data in chunks, applies random masking and similarity thresholding during graph construction, and selects instances based on confidence scores from the trained GAT model. Experiments on 13 diverse datasets demonstrate that GAIS consistently outperforms traditional IS methods in terms of effectiveness, achieving high reduction rates (average 96\%) while maintaining or improving model performance. Although GAIS exhibits slightly higher computational costs, its superior performance in maintaining accuracy with significantly reduced training data makes it a promising approach for graph-based data selection. Code is available at \url{https://github.com/zahiriddin-rustamov/gais}.
\end{abstract}

\begin{IEEEkeywords}
instance selection, graph attention networks, machine learning, data reduction
\end{IEEEkeywords}

\section{Introduction}\label{section:introduction}

In the rapidly evolving landscape of machine learning (ML), the drive towards efficiency is not just a wish but also a necessity, particularly in the context of big data, ML model deployment on edge devices, and the emphasis on the potential of TinyML \cite{warden2019tinyml}. While big data offers rich datasets leading to more accurate models, it also brings high computational demands, longer training times, and increased costs. These challenges are significant in fields requiring rapid model deployment, such as healthcare, finance, and autonomous systems, and in deploying ML models on resource-constrained edge devices. Moreover, many domains struggle with big data due to the costly and time-intensive nature of the data acquisition processes \cite{rodriguez2023deep}. 

One way to tackle these issues is through instance selection (IS), which involves curating training datasets to include only the most informative instances. IS offers dual benefits: reducing computational demands and potentially enhancing model performance by eliminating redundant or noisy data. This is especially crucial in developing TinyML models, where the balance between model size, accuracy, and computational efficiency is critical.

Existing IS methods, such as random sampling, prototype-based methods \cite{garcia2012prototype}, and active learning \cite{fu2013survey}, often fail to capture complex relationships between instances. Hence, we propose a novel IS method using Graph Attention Networks (GATs), which effectively handle graph-structured data by attending to node neighbours. By representing the dataset as a graph and leveraging GATs' attention mechanisms, our method identifies the most informative instances while considering their relationships and overall data structure.
The main contributions of this paper are as follows:
\begin{itemize}
    \item We propose Graph Attention-based Instance Selection (GAIS), which utilizes GATs to identify the most informative instances in the training data based on their local relationships.
    \item We demonstrate GAIS's effectiveness on benchmark datasets, achieving significant dataset size reductions while maintaining model performance, highlighting its superiority over existing IS methods.
\end{itemize}

The paper is structured as follows: Section \ref{section:related_works} presents related work; Section \ref{subsection:problem_def} formulates the problem; Section \ref{section:gais} details the proposed method; Section \ref{section:experimental_setup} describes the experimental setup; Section \ref{section:results} discusses the results; and Section \ref{section:conclusion} concludes the study.

\section{Related Works}\label{section:related_works}

Significant efforts have addressed the challenges of IS and reduction across various domains. Hart's \cite{hart1968condensed} condensed nearest neighbour rule laid the foundation for many subsequent IS algorithms by reducing dataset size while preserving classification integrity. Wilson \cite{wilson1972asymptotic} expanded on this by exploring asymptotic properties of nearest neighbour rules with edited data. Brighton and Mellish \cite{brighton2002advances} advanced IS in instance-based learning, offering insights into optimizing IS algorithms.

\subsection{Traditional IS Methods}

Traditional IS methods have been extensively studied and applied in various domains. Zhang et al. \cite{zhang2013} tackled large dataset challenges, while Yu et al. \cite{yu2013multi} improved speech synthesis via prosodic IS. Fu et al. \cite{fu2013survey} surveyed active learning from an IS perspective. Advancements in Multiple-Instance Learning (MIL) and optimization were contributed by Zhang et al. \cite{zhang2013}, and Peng et al. \cite{peng2014outlier}, introducing IS strategies and outlier detection to enhance learning model efficiency and accuracy.

The versatility of IS techniques was further explored by Lee et al. \cite{lee2014evolutionary} and Klinger et al. \cite{klinger2015instance} through evolutionary algorithms based on fuzzy models and cross-lingual training. Lin et al. \cite{lin2015learning}, Stanovov et al. \cite{stanovov2016self}, and Bicici \cite{biccici2016parfda} expanded IS applications to large-scale data detection, hybrid evolutionary algorithms, and statistical machine translation.

Building on these foundations, Carbonera and Abel introduced the centroid density-based IS algorithm (CDIS) \cite{carbonera2015density, carbonera2016novel}, utilizing density measures to enhance IS. Malhat et al. \cite{malhat2020new} proposed GDIS and EGDIS algorithms, leveraging global density and relevance functions to improve classification accuracy and reduction rates.

The field has continued to evolve, with Ashfaq et al. \cite{ashfaq2017toward} and Villuendas-Rey et al. \cite{villuendas2017simultaneous} introducing fuzziness-based intrusion detection and combined instance and feature selection. More recent studies by Kavrin et al. \cite{kavrin2020bagging}, Yuan et al. \cite{yuan2020multiple}, and DeVries et al. \cite{devries2020instance} focus on bagging-based IS, MIL with multiple-point concepts, and IS for generative adversarial networks (GANs), respectively, indicating a trend towards more sophisticated and integrated approaches.

\subsection{Graph-based IS Methods}

Graph-based methods have advanced IS in ML by leveraging data structure, representing datasets as graphs with instances as nodes and relationships as edges. By exploiting the graph structure, this approach captures complex patterns and relationships, leading to more effective IS.

Several graph-based IS methods have been proposed. Nikolaidis et al. \cite{nikolaidis2012spectral} introduced a spectral graph theory-based instance reduction algorithm. Nguyen et al. \cite{nguyen2017graph} focused on clustering visual instances from image collections, and Wang et al. \cite{wang2018instance} enhanced graph-based semi-supervised learning through IS. Yang et al. \cite{yang2018natural} developed the Natural Neighborhood Graph-based Instance Reduction (NNGIR), which automatically constructs a natural neighborhood graph to divide the training set into noisy, border, and internal instances, eliminating noisy and redundant points without requiring parameter tuning.

Other notable methods include Gabriel Graph Editing (GGE) and Relative Neighborhood Graph Editing (RNGE) \cite{sanchez1997prototype}, which use geometric criteria to connect close points and retain instances near decision boundaries. Hit Miss Networks (HMNs) \cite{marchiori2008hit} represent training sets as graphs where nodes are instances and edges represent nearest neighbor relations, leading to editing algorithms that remove instances based on graph properties. Class Conditional IS \cite{marchiori2009class} builds on HMNs by introducing a scoring function for large margin IS.
Recent studies by Nonaka et al. \cite{nonaka2020graph} and Lou et al. \cite{lou2023asiam} demonstrate the adaptability of graph-based methods in IS, analyzing deep learning outputs using K-nearest neighbor graphs and adopting heterogeneous graph neural networks, respectively.

Despite advancements, traditional IS methods face challenges with scalability, noisy data, and balancing reduction rate with classification accuracy, especially with complex, high-dimensional data \cite{garcia2012prototype, malhat2020new}. Standard graph-based methods may still struggle with complex data and often lack the ability to effectively capture and leverage the varying importance of different neighboring instances, highlighting the need for more adaptive approaches.

GATs present a significant advancement over traditional methods \cite{zaki2023graph}. GATs employ a dynamic attention mechanism that adaptively assigns importance to nodes, allowing for more nuanced handling of complex inter-instance relationships. This approach offers increased robustness against noise in the data. The key innovation of GATs lies in their ability to learn and apply varying weights to different neighboring instances, potentially capturing subtle but important patterns in the data structure. This dynamic weighting mechanism could lead to more informed IS decisions, particularly in datasets with complex or hierarchical relationships between data points.
Building on these advancements, this paper introduces \emph{GAIS}, a novel IS method that integrates GATs to explore how adaptive attention mechanisms can be applied to the challenge of data reduction in ML.

\section{Problem Definition}\label{subsection:problem_def}

The IS problem seeks to select a minimal subset $S \subseteq D$ from a dataset $D = \{(x_i, y_i)\}_{i=1}^N$, where $x_i$ is the feature vector and $y_i$ the corresponding label, such that a ML model trained on $S$ performs comparably to one trained on $D$. Formally, the objective is:
\[
\min_{S \subseteq D} |S|
\]
subject to:
\[
\text{Accuracy}(M(S)) \geq \text{Accuracy}(M(D)) - \Phi
\]
where $|S|$ is the size of $S$, $M(S)$ is the model trained on $S$, $\text{Accuracy}(M(S))$ its accuracy, and $\Phi$ is a tolerance threshold. The goal is to find the smallest $S$ that maintains acceptable performance compared to training on the full dataset $D$.

\section{Graph Attention-based Instance Selection (GAIS)}\label{section:gais}

The proposed GAIS method aims to efficiently reduces dataset size by preserving representative samples using GATs. It processes data in chunks to select informative instances based on confidence scores. The workflow includes data chunking, graph construction, GAT model training, and IS.

\subsection{Data Chunking}
Given a dataset \(\mathcal{D} = \{(\mathbf{x}_i, y_i)\}_{i=1}^N\), GAIS divides it into manageable overlapping chunks \(\mathcal{C}_j\) of window size \(w\) (e.g., 8,000) with overlap \(o\):
\[
\mathcal{C}_j = \{(\mathbf{x}_i, y_i)\}_{i=(j-1)(w-o)+1}^{j(w-o)+o}, \quad j = 1, 2, \dots, \left\lceil\frac{N-o}{w-o}\right\rceil
\]
where \(\left\lceil\cdot\right\rceil\) denotes the ceiling function. The motivation behind chunking the data is to address computational and memory constraints when constructing graphs from large datasets. Chunking allows for localized graph construction within each chunk, speeding up computation. It also handles large datasets incrementally, preventing memory overload. Overlapping chunks maintain continuity between instances across boundaries.

Prior to chunking, we randomly shuffle the \(\mathcal{D}\) to ensure that the chunking process is not dependent on the original order of the data. This step helps to mitigate any potential bias from inherent data ordering and improves the robustness of our method to data permutations.

\subsection{Graph Construction}
For each chunk \(\mathcal{C}_j\), GAIS constructs an undirected graph \(\mathcal{G}_j = (\mathcal{V}_j, \mathcal{E}_j)\) with edges determined by a chosen distance metric (e.g., Euclidean distance). Edge weights \(a_{ij}\) between nodes \(i\) and \(j\) are computed accordingly. To reduce complexity and focus on significant connections, GAIS applies two thresholds:
\begin{itemize}
    \item \textbf{Random Threshold} (\(\theta_r\)): Randomly removes a fraction of edges to introduce randomness and prevent overfitting.
    \item \textbf{Similarity Threshold} (\(\theta_s\)): Removes edges with weights below \(\theta_s\), eliminating weak connections.
\end{itemize}
The resulting adjacency matrix \(\mathbf{A}_j\) is defined as:
\[
\mathbf{A}_j[i, j] =
\begin{cases}
a_{ij}, & \text{if } a_{ij} \geq \theta_s \text{ and } (i, j) \notin \mathcal{R}_j \\
0, & \text{otherwise}
\end{cases}
\]
where \(\mathcal{R}_j\) is the set of randomly removed edges based on \(\theta_r\).

\subsection{GAT Model Architecture}
GAIS employs a GAT as its core component to assign attention weights to edges, capturing the importance of each neighbor. The GAT architecture consists of input, hidden, and output layers.
A single GAT layer updates node features as:
\[
\mathbf{h}_i' = \text{ELU}\left( \sum_{j \in \mathcal{N}(i)} \alpha_{ij} \mathbf{W} \mathbf{h}_j \right)
\]
where \(\mathbf{h}_i'\) and \(\mathbf{h}_i\) are the output and input features of node \(i\), \(\mathcal{N}(i)\) denotes its neighbors, \(\mathbf{W}\) is a learnable weight matrix, and \(\alpha_{ij}\) is the attention coefficient:
\[
\alpha_{ij} = \frac{\exp\left(\text{LeakyReLU}\left(\mathbf{a}^\top [\mathbf{W}\mathbf{h}_i \parallel \mathbf{W}\mathbf{h}_j]\right)\right)}{\sum_{k \in \mathcal{N}(i)} \exp\left(\text{LeakyReLU}\left(\mathbf{a}^\top [\mathbf{W}\mathbf{h}_i \parallel \mathbf{W}\mathbf{h}_k]\right)\right)}
\]
with \(\mathbf{a}\) as a learnable vector and \(\parallel\) denoting concatenation.

Stacking multiple GAT layers captures higher-order relationships. The GAIS model architecture is defined as:
\[
\mathbf{Z} = \text{GATLayer}_2 \left( \text{ELU}\left( \text{GATLayer}_1(\mathbf{X}, \mathbf{A}) \right), \mathbf{A} \right)
\]
where \(\mathbf{Z}\) is the output feature matrix, \(\mathbf{X}\) the input features, and \(\mathbf{A}\) the adjacency matrix. The final output is passed through a softmax function to obtain node representations \(\mathbf{z}_i\) for IS.

Node representations are updated by aggregating weighted neighbor features:
\[
\mathbf{h}_i^{(l)} = \sigma\left( \sum_{j \in \mathcal{N}(i)} \alpha_{ij}^{(l)} \mathbf{W}^{(l)} \mathbf{h}_j^{(l-1)} \right)
\]
where \(\sigma\) is an activation function like ReLU or ELU. This architecture effectively weighs neighbor influences, transforming input features into intermediate representations and output probabilities.

\subsection{Sequential Training}

GAIS trains the GAT model sequentially on data chunks \(\mathcal{C}_j\), initializing weights randomly and updating iteratively per chunk. This approach handles large datasets by focusing on subsets, reducing memory requirements and enabling efficient learning.

The training objective minimizes the weighted negative log-likelihood loss:
\[
\mathcal{L}_{\text{w}} = \frac{1}{|\mathcal{V}_j|} \sum_{i=1}^{|\mathcal{V}_j|} \left( - \sum_{c=1}^C y_{ic} \log(\hat{y}_{ic}) \right) s_i
\]
where \(C\) is the number of classes, \(y_{ic}\) is the true label, \(\hat{y}_{ic}\) is the predicted probability (from softmax of \(\mathbf{z}_i\)), and \(s_i\) is the confidence score for node \(i\). This weighting focuses the model on instances with higher confidence, enhancing learning and performance.

\begin{table*}[!ht]
\centering
\caption{Datasets Overview. $N$, $P$, and $C$ represent the number of instances, features, and classes, respectively.}\label{table:datasets}
\begin{tabular*}{0.8\linewidth}{@{\extracolsep\fill} 
    >{\raggedright\arraybackslash}p{0.11\linewidth}
    >{\raggedright\arraybackslash}p{0.14\linewidth}
    >{\raggedright\arraybackslash}p{0.3\linewidth} 
    >{\centering\arraybackslash}p{0.05\linewidth}
    >{\centering\arraybackslash}p{0.03\linewidth}
    >{\centering\arraybackslash}p{0.03\linewidth}}
\toprule
Dataset Name & Domains & Objective & $N$ & $P$ & $C$ \\
\midrule
Heart Disease & Healthcare & Predict heart disease presence & 303 & 14 & 2 \\
HCC & Oncology & Identify Hepatocellular Carcinoma cases & 615 & 13 & 2 \\
Stroke Prediction & Healthcare & Predict stroke likelihood & 749 & 11 & 2 \\
Diabetes & Healthcare & Predict diabetes onset & 768 & 9 & 2 \\
German Credit & Financial Risk & Classify credit risks & 1,000 & 20 & 2 \\
Chess End-Game & Game Theory & Predict game outcome & 3,196 & 36 & 2 \\
Spam-base & Email Classification & Classify emails as spam or non-spam & 4,601 & 57 & 2 \\
Two-norm & Synthetic Data & Distinguish between two classes & 7,400 & 20 & 2 \\
Ring-norm & Synthetic Data & Classify into one of the rings & 7,400 & 20 & 2 \\
Coil-2000 & Image Recognition & Object recognition and classification & 9,822 & 85 & 2 \\
Nursery & Social Science & Rank applications for nursery schools & 12,960 & 8 & 2 \\
Magic & Physics \& Chemistry & Classify gamma-ray and hadron events & 19,020 & 10 & 2 \\
Fars & Traffic Safety & Classify injury levels suffered in car accidents & 100,968 & 29 & 9 \\
\bottomrule
\end{tabular*}
\end{table*}

\subsection{Instance Selection}
After training on all data chunks, GAIS performs IS by reprocessing each chunk \(\mathcal{C}_j\). For each instance \(i\) in \(\mathcal{C}_j\), the model computes a confidence score \(s_i\):
\[
s_i = \max_c \hat{y}_{ic}
\]
Instances with confidence scores above a predefined threshold \(\theta_c\) are considered informative and included in the final dataset \(\mathcal{D}_{\text{GAIS}}\):
\[
\mathcal{D}_{\text{GAIS}} = \bigcup_{j} \left\{ (\mathbf{x}_i, y_i) \in \mathcal{C}_j \mid s_i \geq \theta_c \right\}
\]
This step allows GAIS to identify key instances that capture the dataset's essential characteristics, reducing size while preserving informative samples.

\subsection{Hyperparameter Tuning Strategy}
We employ Bayesian optimization via the Optuna framework \cite{Akiba2019} to tune the GAT's hyperparameters \(\mathbf{H}^*\), maximizing performance metric \(E\) on a validation set. The hyperparameters include hidden dimension \(h\), number of input and output heads \(n_{\text{in}}\), \(n_{\text{out}}\), dropout rate \(\delta\), distance metric \(d\) and thresholds ($\theta_r$), ($\theta_s$), and ($\theta_c$) for random, similarity, and confidence, respectively.

Bayesian optimization involves modeling the objective function \(f(\mathbf{H})\), representing the model’s performance as a function of hyperparameters, with a surrogate probabilistic model (e.g., Gaussian Process). An acquisition function \(a(\mathbf{H}; \mathcal{D})\), such as Expected Improvement (EI) or Upper Confidence Bound (UCB), guides the selection of the next hyperparameters by balancing exploration and exploitation:
\[
\text{EI}(\mathbf{H}) = \mathbb{E}\left[\max(f(\mathbf{H}) - f(\mathbf{H}^+), 0)\right],
\]
\[
\text{UCB}(\mathbf{H}) = \mu(\mathbf{H}) + \kappa\sigma(\mathbf{H})
\]
where \(\mathbf{H}^+\) is the current best hyperparameter set, \(\kappa\) is a parameter, and \(\sigma(\mathbf{H})\) is the standard deviation at \(\mathbf{H}\). The next hyperparameters are chosen by maximizing the acquisition function:
\[
\mathbf{H}_{\text{next}} = \arg\max_{\mathbf{H}} a(\mathbf{H}; \mathcal{D})
\]
This iterative process updates the surrogate model with new evaluations until convergence, ensuring the GAT model is optimally tuned for each dataset.

\section{Experimental Setup}\label{section:experimental_setup}

This section details the experimental framework used to evaluate the GAIS method.

\subsection{Datasets}

In this study, we used a diverse set of well-known datasets to evaluate the effectiveness of GAIS, summarised in Table \ref{table:datasets}. The datasets cover domains such as healthcare, financial risk assessment, game theory, oncology, and image recognition. This diversity tests the robustness and applicability of our method across different scenarios.

The datasets are split into training (80\%), validation (10\%), and testing (10\%) subsets. The training set was used to apply the proposed GAIS method for IS; the validation set optimized hyperparameters for GAIS and GAT; the testing set evaluated ML model performance on reduced datasets. We controlled randomness in splitting for reproducibility. All features were retained; categorical features were label encoded, numerical features scaled to [0,1], and target columns standardized for consistency.

\subsection{Graph Construction}

To apply GAIS, we converted datasets into graph representations using three distance metrics: Manhattan, Euclidean, and cosine similarity, capturing different aspects of instance similarity. For a dataset $X$ with $n$ instances, we computed pairwise distances $D \in \mathbb{R}^{n \times n}$. We transformed distances into similarity scores using:
\[
S_{ij} = \begin{cases}
1 - D_{ij}, & \text{for cosine similarity} \\
1 / (1 + D_{ij}), & \text{for Manhattan and Euclidean}
\end{cases}
\]
A similarity threshold (\(\theta_s\)) was applied to create edges only between instances with similarity above this threshold, ensuring connections among the most similar instances.

\subsection{Machine Learning Models}
We evaluated GAIS using five ML models: naive bayes (NB), random forest (RF), logistic regression (LR), eXtreme gradient boosting (XGB), and k-nearest neighbours (KNN), chosen for their diverse learning paradigms and ability to handle various data types and complexities.

\begin{table*}[t]
\centering
\caption{Performance Metrics Over 13 Datasets Based on Reduced and Full Training Data.}\label{table:results}
\begin{tabular*}{0.93\linewidth}{@{\extracolsep\fill}
    >{\raggedright\arraybackslash}p{0.11\linewidth}
    >{\centering\arraybackslash}p{0.05\linewidth}|
    >{\centering\arraybackslash}p{0.05\linewidth}
    >{\centering\arraybackslash}p{0.05\linewidth}
    >{\centering\arraybackslash}p{0.05\linewidth}
    >{\centering\arraybackslash}p{0.05\linewidth}|
    >{\centering\arraybackslash}p{0.05\linewidth}
    >{\centering\arraybackslash}p{0.05\linewidth}
    >{\centering\arraybackslash}p{0.05\linewidth}
    >{\centering\arraybackslash}p{0.05\linewidth}|
    >{\centering\arraybackslash}p{0.05\linewidth}
    >{\centering\arraybackslash}p{0.05\linewidth}}
\toprule
 &  & \multicolumn{4}{c|}{\textbf{Reduced Data}} & \multicolumn{4}{c|}{\textbf{Original Data}} &  &  \\
\textbf{Dataset} & \textbf{$TR_A$} & \textbf{$AC$} & \textbf{$PR$} & \textbf{$RE$} & \textbf{$F1$} & \textbf{$AC$} & \textbf{$PR$} & \textbf{$RE$} & \textbf{$F1$} & \textbf{$R$} & \textbf{$E$} \\
\midrule
   Heart Disease & 244 & 0.9355 & 1.0000 & 0.8824 & 0.9375 & 0.9032 & 0.9375 & 0.8824 & 0.9091 & 0.9549 & 0.8933 \\
   HCC & 497 & 0.9516 & 1.0000 & 0.6250 & 0.7692 & 0.9839 & 0.8889 & 1.0000 & 0.9412 & 0.9799 & 0.9324 \\
   Stroke Prediction & 606 & 0.8133 & 0.7200 & 0.7200 & 0.7200 & 0.7733 & 0.6538 & 0.6800 & 0.6667 & 0.9884 & 0.8862 \\
   Diabetes & 621 & 0.7273 & 0.6250 & 0.5556 & 0.5882 & 0.7013 & 0.6250 & 0.3704 & 0.4651 & 0.9871 & 0.7179 \\
   German Credit & 810 & 0.8000 & 0.7778 & 0.4667 & 0.5833 & 0.7600 & 0.6000 & 0.6000 & 0.6000 & 0.9580 & 0.7664 \\
   Chess End-Game & 2588 & 0.9469 & 0.9573 & 0.9401 & 0.9486 & 0.9906 & 0.9824 & 1.0000 & 0.9911 & 0.9629 & 0.9118 \\
   Spam-base & 3726 & 0.8980 & 0.9592 & 0.7747 & 0.8571 & 0.9501 & 0.9492 & 0.9231 & 0.9359 & 0.9868 & 0.9655 \\
   Ring-norm & 5994 & 0.9828 & 0.9661 & 0.9893 & 0.9775 & 0.9770 & 0.9661 & 0.9893 & 0.9775 & 0.9828 & 0.9655 \\
   Two-norm & 5994 & 0.9730 & 0.9534 & 0.9946 & 0.9735 & 0.9703 & 0.9780 & 0.9622 & 0.9700 & 0.9863 & 0.8039 \\
   Coil-2000 & 7955 & 0.9156 & 0.1667 & 0.1017 & 0.1263 & 0.9390 & 0.0000 & 0.0000 & 0.0000 & 0.9901 & 0.9064 \\
   Nursery & 10495 & 0.9460 & 0.9248 & 0.8533 & 0.8804 & 0.9992 & 0.9994 & 0.9994 & 0.9994 & 0.9334 & 0.8830 \\
   Magic & 15406 & 0.8617 & 0.8652 & 0.9319 & 0.8973 & 0.8906 & 0.8891 & 0.9497 & 0.9184 & 0.9627 & 0.8296 \\
   Fars & 81783 & 0.7784 & 0.5371 & 0.5113 & 0.5182 & 0.8015 & 0.6043 & 0.5552 & 0.5606 & 0.9278 & 0.7221 \\
\midrule
   \textbf{Average} &  & \textbf{0.8869} & \textbf{0.8046} & \textbf{0.7192} & \textbf{0.7525} & \textbf{0.8954} & \textbf{0.7749} & \textbf{0.7624} & \textbf{0.7642} & \textbf{0.9693} & \textbf{0.8599} \\
\bottomrule
\end{tabular*}
\end{table*}

\subsection{Evaluation Metrics}
We assess the performance of the GAIS method using a range of standard evaluation metrics. Accuracy (AC) measures the proportion of correctly classified instances out of the total instances, calculated as:
\[
AC = ({TP + TN})/({TP + TN + FP + FN})
\]
where $TP$, $TN$, $FP$, and $FN$ denote true positives, true negatives, false positives, and false negatives, respectively.
Precision (PR) represents the proportion of true positive instances among the instances predicted as positive:
\[
PR = {TP}/({TP + FP})
\]
Recall (RE) is the proportion of true positive instances among the actual positive instances:
\[
RE = {TP}/({TP + FN})
\]
F1-score (F1) is the harmonic mean of precision and recall, providing a balanced measure of a model's performance, calculated as:
\[
F1 = 2 \times ({PR \times RE})/({PR + RE})
\]

To evaluate the balance between dataset reduction and accuracy, we employ the Effectiveness metric \cite{malhat2020new}. The Reduction Rate ($R$) quantifies the proportion of instances removed:
\[
R = 1 - {|S|}/{|D|}
\]
where $|S|$ and $|D|$ are the sizes of the reduced and the original dataset, respectively. The Effectiveness ($E$) is calculated as:
\[
E = AC \times R
\]
We also measure the total computation time ($T_{IS}$) of the GAIS method, including preprocessing, GAT training, thresholding, and ML model training.

Experiments were conducted on a Windows 10 system with an Intel Core i7-12700 CPU (12 cores, 20 threads), 128 GB RAM, and an NVIDIA GeForce RTX 4090 GPU (with 24 GB memory). GAT was implemented using PyTorch 2.1.2 and PyTorch Geometric 2.4.0.

\section{Results \& Discussions}\label{section:results}

This section presents the experimental results obtained by evaluating the performance of the proposed GAIS method on various datasets. Table \ref{table:results} summarizes the results, comparing the performance metrics achieved on the reduced datasets obtained using GAIS with those obtained on the original datasets. The table includes metrics such as the number of original training instances ($TR_A$), accuracy ($AC$), precision ($PR$), recall ($RE$), F1 score ($F1$), reduction rate ($R$), and effectiveness ($E$). From the results presented, several insights can be drawn:
\begin{itemize}
    \item The proposed GAIS method achieves high reduction rates ($R$) across all datasets, with an average reduction rate of 0.9693. This indicates that GAIS effectively reduces the size of the datasets while maintaining comparable performance to the original datasets.
    \item The accuracy ($AC$) obtained on the reduced datasets is close to that of the original datasets, with an average difference of only 0.0085. This suggests that GAIS is able to select informative instances that contribute to the learning process.
    \item The effectiveness ($E$) metric, which combines accuracy and reduction rate, shows that GAIS achieves a good balance between dataset reduction and performance, with an average effectiveness of 0.8599.
    \item While the precision ($PR$), recall ($RE$), and F1 score ($F1$) vary across datasets, the average values for these metrics on the reduced datasets are comparable to those obtained on the original datasets, indicating that GAIS maintains the overall performance of the models.
    \item In some cases, the reduced datasets obtained using GAIS perform better than the original datasets. For example, on the Heart Disease dataset, the accuracy on the reduced dataset (0.9355) is higher than that on the original dataset (0.9032). Similarly, on the Diabetes dataset, the precision, recall, and F1 score are all higher on the reduced dataset compared to the original dataset. These instances demonstrate that GAIS can potentially remove noisy or less informative instances, leading to improved model performance.
    \item There are also cases where the performance on the reduced datasets is slightly lower than that on the original datasets. For instance, on the Nursery dataset, the accuracy on the reduced dataset (0.9460) is lower than that on the original dataset (0.9992). However, it is important to consider the trade-off between dataset reduction and performance. In this case, GAIS achieves a high reduction rate of 0.9334 while still maintaining a relatively high accuracy.
\end{itemize}
The results indicate that leveraging GATs for IS can effectively reduce dataset size while maintaining model performance. The high reduction rates achieved, coupled with the preservation of performance metrics, suggest that the adaptive attention mechanism contributes to more informed IS.

\begin{figure*}[!ht]
    \centering
    \includegraphics[width=0.8\linewidth]{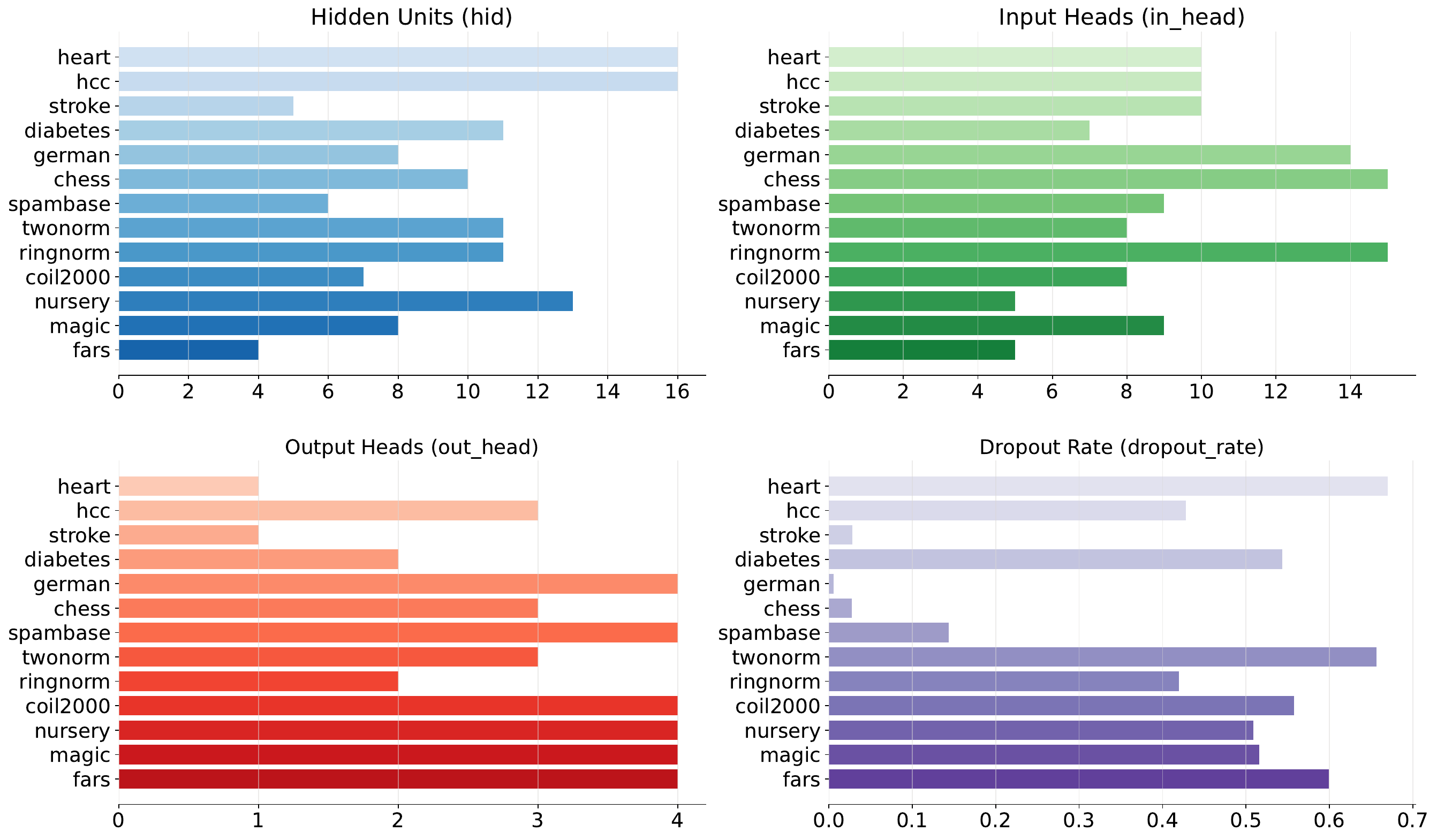}
    \caption{GAT Hyperparameter Distribution Over the Datasets.}
    \label{fig:gat_parameters}
\end{figure*}

\begin{table*}
\centering
\caption{Comparative Performance Analysis of IS Methods Based on Effectiveness Metric.}\label{table:comparative-e}
\begin{tabular*}{0.8\linewidth}{@{\extracolsep\fill}
    >{\raggedright\arraybackslash}p{0.12\linewidth}
    >{\centering\arraybackslash}p{0.06\linewidth}
    >{\centering\arraybackslash}p{0.06\linewidth}
    >{\centering\arraybackslash}p{0.06\linewidth}
    >{\centering\arraybackslash}p{0.06\linewidth}
    >{\centering\arraybackslash}p{0.06\linewidth}
    >{\centering\arraybackslash}p{0.06\linewidth}
    >{\centering\arraybackslash}p{0.06\linewidth}
    >{\centering\arraybackslash}p{0.06\linewidth}}
\toprule
\textbf{Dataset} & \textbf{CNN} & \textbf{DROP3} & \textbf{ENN} & \textbf{IB3} & \textbf{ICF} & \textbf{LDIS} & \textbf{RMHC} & \textbf{GAIS} \\
\midrule
Heart Disease & 0.460 & 0.590 & 0.645 & 0.168 & 0.553 & 0.550 & 0.500 & \textbf{0.893} \\
HCC & 0.839 & - & 0.067 & 0.750 & 0.851 & 0.610 & 0.428 & \textbf{0.932} \\
Stroke Prediction & 0.375 & 0.651 & 0.213 & 0.482 & 0.473 & 0.644 & 0.379 & \textbf{0.804} \\
Diabetes & 0.360 & 0.641 & 0.185 & 0.381 & 0.487 & 0.652 & 0.388 & \textbf{0.718} \\
German Credit & 0.375 & 0.711 & 0.224 & 0.202 & 0.335 & 0.660 & 0.385 & \textbf{0.766} \\
Chess End-Game & 0.778 & 0.588 & 0.102 & 0.618 & 0.844 & 0.753 & 0.496 & \textbf{0.912} \\
Spam-base & 0.716 & 0.832 & 0.093 & 0.602 & 0.779 & 0.756 & 0.447 & \textbf{0.886} \\
Ring-norm & 0.717 & - & 0.250 & 0.749 & 0.784 & 0.405 & 0.486 & \textbf{0.966} \\
Two-norm & 0.806 & 0.936 & 0.324 & 0.826 & 0.915 & 0.824 & 0.492 & \textbf{0.960} \\
Coil-2000 & 0.707 & 0.303 & 0.145 & 0.809 & 0.739 & 0.838 & 0.463 & \textbf{0.906} \\
Nursery & 0.703 & - & 0.093 & 0.697 & 0.370 & 0.550 & 0.504 & \textbf{0.883} \\
Magic & 0.562 & 0.720 & 0.169 & 0.639 & 0.624 & 0.731 & 0.445 & \textbf{0.830} \\
Fars & 0.465 & - & - & - & 0.528 & 0.516 & 0.400 & \textbf{0.722} \\
\midrule
\textbf{Average} & 0.605 & 0.664 & 0.209 & 0.577 & 0.637 & 0.653 & 0.447 & \textbf{0.860} \\
\bottomrule
\end{tabular*}
\end{table*}

Figure \ref{fig:gat_parameters} illustrates the optimal GAT hyperparameters across datasets, revealing several insights:

\begin{itemize}
    \item \textbf{Hidden Units:} Optimal values range from 4 to 16, with most datasets between 7 and 11. Higher units (e.g., 16 in "heart" and "hcc") suggest more complex patterns, while lower units (e.g., 4 in "fars" and 5 in "stroke") indicate simpler architectures suffice.
    \item \textbf{Input Heads:} Optimal values range from 5 to 15, commonly between 8 and 10. Higher input heads (e.g., 15 in "chess" and "ringnorm") are beneficial for capturing complex relationships, whereas lower input heads (e.g., 5 in "nursery" and "fars") are adequate for simpler datasets.
    \item \textbf{Output Heads:} Less variation is observed, with optimal values from 1 to 4, mostly 3 or 4. Multiple output representations generally enhance performance, but datasets like "heart" and "stroke" perform best with a single output head.
    \item \textbf{Dropout Rate:} Optimal dropout rates vary widely from 0.005 to 0.67, indicating dataset-dependent regularization needs. Minimal dropout (e.g., 0.005 in "german") suggests less overfitting risk, while higher dropout (e.g., 0.67 in "heart") indicates a need for stronger regularization.
\end{itemize}

\begin{figure*}[!ht]
    \centering
    \begin{subfigure}[b]{0.46\linewidth}
        \centering
        \includegraphics[width=\linewidth]{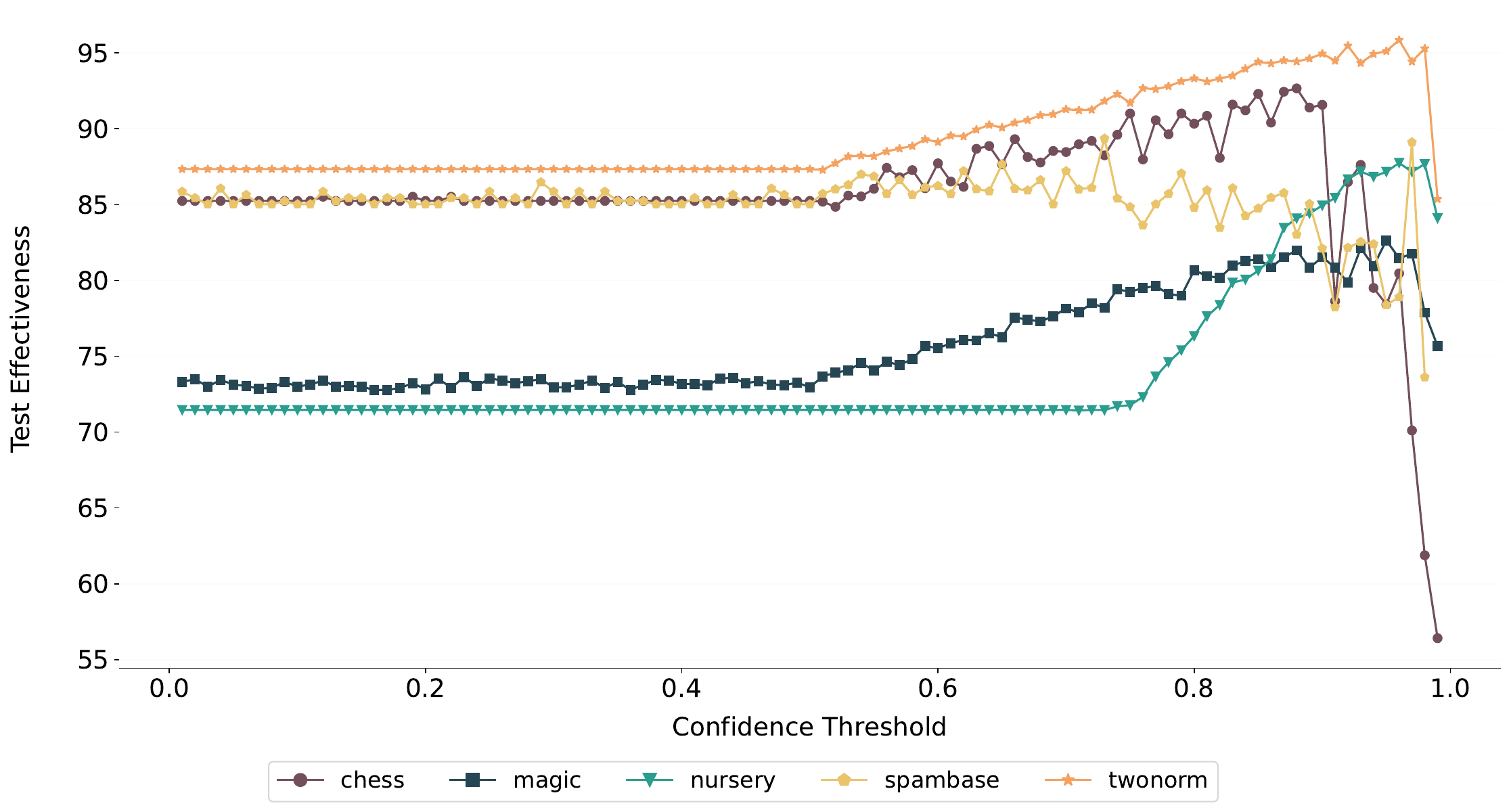}
        \caption{Confidence Threshold}
        \label{fig:confidence_threshold}
    \end{subfigure}
    \hspace{0.01\linewidth}
    \begin{subfigure}[b]{0.46\linewidth}
        \centering
        \includegraphics[width=\linewidth]{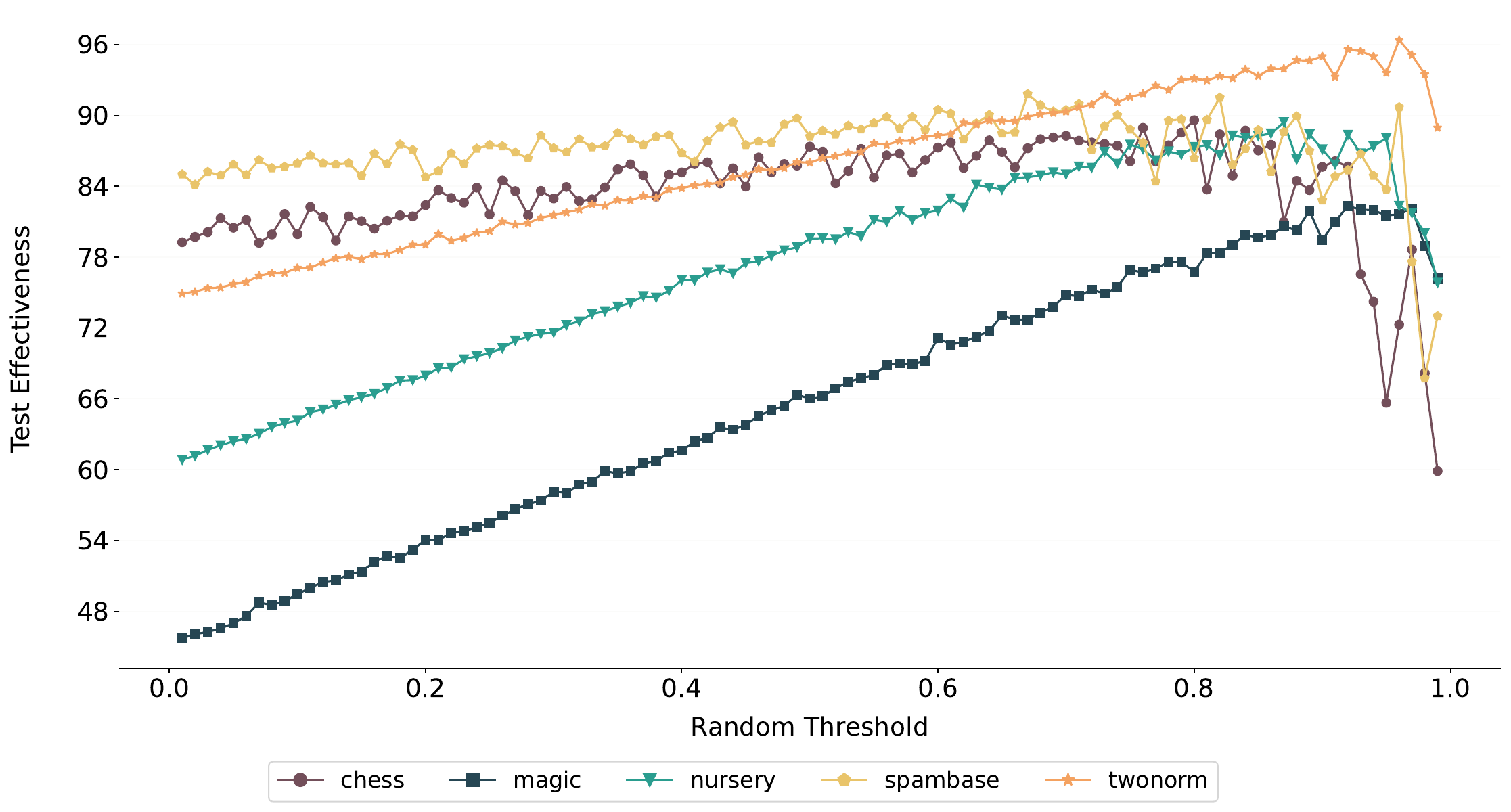}
        \caption{Random Threshold}
        \label{fig:random_threshold}
    \end{subfigure}
    \caption{The Effect of GAIS Hyperparameters on Effectiveness Across Five Datasets.}
    \label{fig:gais_parameters}
\end{figure*}

\begin{table*}
\centering
\caption{Comparative Performance Analysis of IS Methods Based on the IS Time.}\label{table:comparative-tis}
\begin{tabular*}{0.8\linewidth}{@{\extracolsep\fill}
    >{\raggedright\arraybackslash}p{0.12\linewidth}
    >{\centering\arraybackslash}p{0.06\linewidth}
    >{\centering\arraybackslash}p{0.06\linewidth}
    >{\centering\arraybackslash}p{0.06\linewidth}
    >{\centering\arraybackslash}p{0.06\linewidth}
    >{\centering\arraybackslash}p{0.06\linewidth}
    >{\centering\arraybackslash}p{0.06\linewidth}
    >{\centering\arraybackslash}p{0.06\linewidth}
    >{\centering\arraybackslash}p{0.06\linewidth}}
\toprule
\textbf{Dataset} & \textbf{CNN} & \textbf{DROP3} & \textbf{ENN} & \textbf{IB3} & \textbf{ICF} & \textbf{LDIS} & \textbf{RMHC} & \textbf{GAIS} \\
\midrule
Heart Disease & 0.689 & 0.309 & \textbf{0.003} & 0.731 & 1.195 & 0.006 & 0.175 & 17.4431 \\
HCC & 1.415 & - & \textbf{0.005} & 2.492 & 1.614 & \textbf{0.005} & 0.121 & 7.179 \\
Stroke Prediction & 3.219 & 0.766 & \textbf{0.006} & 1.007 & 1.304 & \textbf{0.006} & 0.274 & 15.410 \\
Diabetes & 1.729 & 0.559 & \textbf{0.006} & 2.415 & 3.601 & 0.007 & 0.018 & 7.241 \\
German Credit & 13.571 & 10.594 & \textbf{0.026} & 11.634 & 15.046 & 0.033 & 0.165 & 9.080 \\
Chess End-Game & 75.336 & 24.462 & 0.104 & 11.883 & 26.673 & \textbf{0.080} & 0.811 & 19.927 \\
Spam-base & 100.473 & 3.981 & 0.041 & 72.207 & 29.926 & \textbf{0.035} & 1.519 & 17.456 \\
Ring-norm & 178.647 & - & 0.059 & 68.873 & 34.232 & \textbf{0.046} & 0.946 & 79.585 \\
Two-norm & 264.087 & 6.505 & \textbf{0.043} & 62.518 & 26.739 & 0.206 & 0.457 & 44.670 \\
Coil-2000 & 239.117 & 6.993 & 0.126 & 58.203 & 69.286 & \textbf{0.106} & 0.827 & 43.727 \\
Nursery & 744.851 & - & 0.189 & 100.068 & 4.222 & \textbf{0.177} & 3.487 & 48.376 \\
Magic & 729.109 & 23.478 & 0.315 & 219.844 & 270.974 & \textbf{0.298} & 6.495 & 205.855 \\
Fars & 2306.831 & - & - & - & 2578.844 & \textbf{2.380} & 63.949 & 387.421 \\
\midrule
\textbf{Average} & 358.390 & 8.627 & 0.077 & 50.990 & 235.666 & 0.260 & 6.096 & 69.490 \\
\bottomrule
\end{tabular*}
\end{table*}

Figure \ref{fig:gais_parameters} illustrates the effect of the confidence threshold (\(\theta_c\)) and the random threshold (\(\theta_r\)) on test effectiveness across five datasets. The confidence threshold in GAIS filters confidence scores from the GAT model to select the most representative instances.

For \(\theta_c\), the general trend shows that at lower values, test effectiveness remains relatively constant. As \(\theta_c\) increases (particularly around 0.6–0.7 and above), test effectiveness rises, suggesting that higher thresholds allow GAIS to select more informative instances, enhancing performance. However, when \(\theta_c\) reaches very high values (e.g., 0.95), test effectiveness drops significantly, likely due to selecting too few instances to capture underlying data patterns.

The random threshold \(\theta_r\) is used to randomly mask training data during GAT model training. Its impact varies across datasets. In some datasets like \textit{magic} and \textit{nursery}, test effectiveness steadily increases with higher \(\theta_r\), indicating that masking more training data helps the GAT model learn more robust representations. In others like \textit{chess} and \textit{spambase}, the increase is marginal. Similar to \(\theta_c\), when \(\theta_r\) is very high (e.g., 0.95), test effectiveness dips across all datasets due to excessive masking leaving insufficient information for learning.

These findings emphasize the importance of carefully tuning \(\theta_c\) and \(\theta_r\) based on dataset characteristics, balancing between masking training data and preserving enough information for effective learning.

The results presented in Tables \ref{table:comparative-e} and \ref{table:comparative-tis} demonstrate the superior performance of GAIS across all datasets compared to well-known IS methods: CNN \cite{hart1968condensed}, DROP3 \cite{DRandall2000}, ENN \cite{wilson1972asymptotic}, IB3 \cite{Aha1991}, ICF \cite{brighton2002advances}, LDIS \cite{carbonera2015density}, and RMHC \cite{Skalak1994}. We also wanted to include graph-based methods, GGE and RNGE, in the comparison, but they were found to be extremely slow and computationally expensive, making them infeasible for practical use.
Table \ref{table:comparative-e} shows the effectiveness ($E$) of various IS methods. The GAIS method consistently achieves the highest effectiveness scores, indicating its ability to select informative and representative instances while maintaining high classification accuracy.
However, as shown in Table \ref{table:comparative-tis}, it is important to note that GAIS's IS time ($T_{IS}$) is slightly higher than some of the other methods. This can be attributed to the fact that GAIS involves preprocessing the data, training the GAT model, tuning the confidence threshold, and training the ML model, which are computationally more expensive than traditional IS techniques.
Despite the longer processing time, the significant improvement in effectiveness demonstrates that GAIS provides a valuable trade-off between computational cost and performance in IS tasks.

\section{Conclusion}\label{section:conclusion}

The proposed GAIS method emerges as a novel and effective approach to IS in ML. Its ability to adapt to different datasets, combined with its innovative use of attention mechanisms, sets it apart from traditional methods. The results across the various datasets not only validate the method’s effectiveness but also highlight its potential to advance IS, particularly in complex and diverse data environments. The proposed method shows promising results in various domains, particularly in maintaining or improving performance with significantly reduced training data. However, there are variations in precision across different datasets, indicating the need for method refinement in certain scenarios. The high effectiveness scores across the board highlight the method’s potential in practical applications where reducing computational resources without sacrificing performance is crucial. These results advocate for further investigation and application of the GAIS method, particularly in environments where computational resources are limited or data acquisition is costly.

While GAIS demonstrates significant adaptability and effectiveness across various datasets, it has limitations. The method's performance depends on resource-intensive hyperparameter tuning, and its scalability is challenged by the computational demands of GAT when dealing with extremely large datasets. Furthermore, GAIS's effectiveness with highly imbalanced datasets and its interpretability in complex scenarios require further exploration.

Future enhancements for the GAIS method could explore several avenues. Algorithmic advancements might focus on expanding network architectures and incorporating recent developments in attention mechanisms. To address scalability challenges, particularly for larger datasets, future research could explore more efficient graph construction techniques and investigate alternative IS methods that do not rely on deep learning approaches. The method's robustness could potentially be improved by integrating data augmentation, transfer learning, and adversarial testing techniques. To enhance the method's applicability in real-world scenarios, future work could emphasize model interpretability and evaluate the method's performance in practical settings.

\bibliographystyle{IEEEtran}

\end{document}